\documentclass{article}
\pdfoutput=1

\usepackage[final]{neurips_2019}

\usepackage[utf8]{inputenc} 
\usepackage[T1]{fontenc}    
\usepackage{hyperref}       
\usepackage{url}            
\usepackage{booktabs}       
\usepackage{amsfonts}       
\usepackage{nicefrac}       
\usepackage{microtype}      
\usepackage{natbib}
\usepackage{makecell}
\usepackage{enumitem}
\usepackage{pdfpages}
\usepackage{standalone}
\usepackage{pifont}
\usepackage{enumitem}
\newcommand{\xmark}{\ding{55}}%

\usepackage{graphicx}
\usepackage{soul}
\usepackage{float}
\usepackage{placeins}
\usepackage{textcomp}
\graphicspath{ {./images/} }

\title{Causal datasheet: An approximate guide to practically assess Bayesian networks in the real world}

\author{%
  Bradley Butcher\thanks{Equal Contribution} \\
  University of Sussex\\
  \texttt{b.butcher@sussex.ac.uk} \\
  \And
  Vincent S. Huang\footnotemark[1] \\
  Surgo Foundation \\
  \texttt{vincenthuang@surgofoundation.org} \\
  \And
  Jeremy Reffin \\
  University of Sussex \\
  \texttt{j.p.reffin@sussex.ac.uk} \\
  \And
  Sema K. Sgaier\thanks{Additional affiliations: Harvard T.H. Chan School of Public Health; University of Washington}\\
  Surgo Foundation \\
  \texttt{semasgaier@surgofoundation.org} \\
  \And
  Grace Charles \\
  Surgo Foundation \\
  \texttt{gracecharles@surgofoundation.org} \\
  \And
  Novi Quadrianto\thanks{Corresponding author}\\
  University of Sussex \\
  \texttt{n.quadrianto@sussex.ac.uk} \\
}
\begin{document}

\maketitle

\begin{abstract}
In solving real-world problems like changing healthcare-seeking behaviors, designing interventions to improve downstream outcomes requires an understanding of the causal links within the system. Causal Bayesian Networks (BN) have been proposed as one such powerful method. In real-world applications, however, confidence in the results of BNs are often moderate at best. This is due in part to the inability to validate against some ground truth, as the DAG is not available. This is especially problematic if the learned DAG conflicts with pre-existing domain doctrine. At the policy level, one must justify insights generated by such analysis, preferably accompanying them with uncertainty estimation. Here we propose a causal extension to the datasheet concept proposed by Gebru et al \citeyearpar{Gebru} to include approximate BN performance expectations for any given dataset. To generate the results for a prototype Causal Datasheet,  we constructed over 30,000 synthetic datasets with properties mirroring characteristics of real data. We then recorded the results given by state-of-the-art structure learning algorithms. These results were used to populate the Causal Datasheet, and recommendations were automatically generated dependent on expected performance. As a proof of concept, we used our Causal Datasheet Generation Tool (CDG-T) to assign expected performance expectations to a maternal health survey we conducted in Uttar Pradesh, India. 
\end{abstract}

\section{Introduction}
Despite causal Bayesian Network's (BN) many offerings, key among which the ability to ask what-if intervention simulations, we have not seen a wide adoption in real-world problems that demand the understanding of interventional outcomes such as private sector marketing design and public health interventions \citep{Arora}. We have found that validating the structure, parameterization, predictive accuracy, and generalizability of BN is subject to considerable debate and interpretation when applied to data with real-world complexity. Our inability to estimate uncertainty in structure learning algorithm performance for specific datasets can call entire models into question. Generally, practitioners must resort to domain expertise to validate model structure\citep{Moglia, Lewis}. This makes BN model results especially difficult to defend when they contradict previous domain beliefs or doctrines. Thus, study results are often presented as a proof-of-concept of BN techniques to show that the method can recover insights already known rather than as an actionable model for discovery, change, or intervention \citep{Lewis,Moglia,Raquejo}.

The problem of not knowing how well machine learning algorithms will perform in real-world conditions is not restricted to causal inference and has been subject to a broader debate.  One proposed solution is adopting the standard practice of constructing and accompanying any given dataset with a full description of the data, its collection context, operating characteristics (i.e., the characteristics of the data to which a machine learning algorithms is applied), and test results (i.e., the expected performance of the machine learning algorithm) \citep{Gebru}. Measuring the expected causal inference performance for any given dataset is, however, not straightforward. First, it is not clear what performance metrics should be used to measure BN algorithms' ability to capture the ground truth causal structure when the ground truth is unknown. In addition, such data may not include the appropriate variables to establish causal or interventional sufficiency \citep{peters2017elements,Spirtes, Kleinberg, Pearl:2009:CMR:1642718}, can have incomplete observations, and may be imbalanced. Perhaps due to the data challenges mentioned above, the evaluation of novel BN algorithms has been largely based on standard synthetic datasets such as ALARM, HEPAR, and others\citep{scutari2009learning,beinlich1989alarm}, which can have vastly different characteristics compared to real-world data at hand. One suggested method for ranking algorithms' performance is to assume the intersection of the structures found by a collection of algorithms as the partial ground truth\citep{pmlr-v84-viinikka18a} but this is neglecting an increasing complexity of dataset when the whole given dataset is considered, nor does it provide a sense of data requirements (e.g., sample size) for different algorithms, which is crucial for data collection design. We face the following quandary: with real-world data we lack the ground truth against which to evaluate the modeling algorithms, and with synthetic data we lack the complexity and limitations that are typically imposed in real-world circumstances \citep{Gentzel}.

Here we propose an approach to attach a causal extension to such datasheet to describe expected causal inference algorithm performance. With it, we wish to empower practitioners to estimate uncertainty levels around the causal structures learnt under the typical contexts and constraints applicable to their analytical problem of interest. We provide two example Causal Datasheets: one generated based on characteristics of a dataset in the global development domain (a survey that we administered in Uttar Pradesh, India), and another based on the characteristics of the well-known ALARM dataset \citep{beinlich1989alarm}. 

\section{Causal Datasheet Generation Tool (CDG-T) Methodology}

Our approach is building a tool to evaluate BN structural learning algorithms on synthetic datasets with known ground truths and with data characteristics mimicking those of real-world datasets. Our starting point was to review a spectrum of real-world datasets; given the authors' own background, we chose global health as our first test domain. We developed a set of characterizing attributes of datasets and determined their typical ranges from surveys and program type data commonly available in global health. Based on these values, an assumed varying set of network structural types, and randomly generated ground truth causal structures, we produced 32,400 synthetic datasets with attributes spanning the range that would mimic real-world datasets. From these synthetic datasets, we learnt a large set of models, applying different structural learning algorithms. We then measured precision and recall of the learned structures, constructing a large lookup table of performances that can extend the datasheet of any given real-world data with similar characterizing attributes. We call this suite of synthetic data generation and performance evaluation tools the Causal Datasheet Generation Tool (CDG-T). 

\subsection{Dataset Characteristics}

To study the variability of structural learning performance with different synthetic data properties, we defined and systematically varied two classes of dataset characteristics: observable and non-observable (Table \ref{tab:Table_observability}). Observable characteristics are those which the designer of the dataset has control and can be easily calculated (e.g., sample size and number of variables). Non-observable characteristics are properties of the underlying truth (e.g., complexity, type of structure, or imbalance). Non-observable characteristics can be estimated, but doing so introduces modeling assumptions. When evaluating a real-world dataset in practice, one could look up a Causal Datasheet with corresponding observable characteristics, to estimate performance uncertainty from the unobservable characteristics.

\begin{table}[h]
\centering
\scalebox{0.9}{
\begin{tabular}{|c|c|c|}
\hline
\textbf{Characteristic}            & \textbf{Observable}  & \textbf{Selected Values} \\ \hline
Samples                   & \checkmark  & \makecell{1,000; 2,500; 5,000; \\ 10,000; 25,000; 50,000}        \\ \hline
Variables                 & \checkmark   & 10, 25, 50, 75, 100           \\ \hline
Alpha (Imbalance)         & \xmark     & 1, 10, 50, 100        \\ \hline
\makecell{Complexity \\ (Median Edges)} & \xmark  & 1, 2, 3             \\ \hline
Structure Type            & \xmark   & \makecell{Forest Fire, \\ Preferential Attachment, \\ IC-DAG}       \\ \hline
Average Variable Levels   & \checkmark  & 3, 4, 5             \\ \hline
\end{tabular}}
\caption{A table on the observability of the properties of BNs (and their associated datasets), as well as the operating characteristics used during the creation of the evaluation of each structure learning algorithm. Each combination of values was used with 10 different seeds resulting in a total of 32,400 datasets, with corresponding true causal structures in the form of BNs.
}
\label{tab:Table_observability}
\end{table}

To best cover the likely spectrum of operating characteristics of global health datasets, a wide range of Directed Acyclic Graphs (DAG), parameter, and dataset property values were used. To determine what represented 'typical' operating characteristics, we conducted a broad literature review of the current usage of BN in global health. We found that typical experiments ranged from fewer than 1,000 samples (e.g., small randomized trials) to greater than 100,000 samples (e.g., electronic health records) and generally used between 10-100 variables. While the evaluation is currently limited to what falls within the confines of typical global health datasets, we intend to extend this to cover a wider range of problems. In addition, we acknowledge that we have not yet addressed real-world data problems such as missing values and mixed continuous and categorical data. Future iterations of Causal Datasheet will address these issues.

\subsection{Structure Learning Algorithms}

Results for the CDG-T look-up table were generated using three structure learning algorithms: a score-based method (FGES), a constraint-based method (PC), and a hybrid approaches (MMHC) \citep{glover, Pearl,  Spirtes, Chickering2002}. The score and hybrid-based approaches both used the BDeu score \citep{buntine}. 

\subsection{Metrics}
To empirically evaluate structure learning methods at different operating characteristics, we measured the precision and recall of the learned structure with respect to the ground truth structure. This allowed us to separate errors into learning false edges versus not identifying true edges, as opposed to quantifying aggregated structural distance measures (e.g., Structural Hamming Distance \citep{de2009comparison}). This can be important during evaluation, as a practitioner may favor precision or recall as the two are often a trade-off. 
We also developed a metric to measure the \textit{proportion of correct interventional odds ratios} (PCOR) to quantify the impact of different learned structures on the interventional odds ratios. The measure was designed to provide an answer the question to practitioners: how trustworthy should these interventional odds ratios be with my dataset?

Due to variation in importance of interventions versus outcomes, we allow the user to specify a threshold of PCOR. The recommendations in the Causal Datasheet will then be based off whether this threshold was met. Ultimately, because PCOR relies on both the structure and parameters of the network being learnt sufficiently well, these do not need to be individually assessed.

\section{Results}

\subsection{Causal Datasheet example: Global Development Dataset}
As a proof of concept, we generated a Causal Datasheet for global development dataset we administered in Uttar Pradesh, India in 2016 (Appendix A). This survey sought to quantify household reproductive, maternal, neonatal, and child health (RMNCH) journeys and to understand the drivers of various RMNCH behaviors. In all, we surveyed over 5,000 women on various RMNCH behaviors and outcomes. From this survey, we identified a total of 18 variables representing critical causal drivers of RMNCH outcomes and behaviors such as birth delivery locations and early breastfeeding initiation. We were interested in understanding which interventions might be most important for different health outcomes. While it was possible to use our dataset to generate DAGs, we could not validate their structures, nor could we assign confidence to graphs generated using different structural learning algorithms. Using survey dataset characteristics, we generated synthetic dataset experiments with similar properties (number of variables = 18, average number of levels = 3, sample size = 6,000) and then attached an expected BN algorithm skeleton, V-structure, and interventional odds ratio performance score to its datasheet for different structure learning algorithms. Because we were most interested in understanding the effects of different interventions on outcomes, we set our PCOR threshold at 0.80. 
The outputs from the CDG-T provided a number of key insights for this dataset. We learned that 1) while all structure learning algorithms could achieve nearly perfect skeleton precision and high skeleton recall, the FGES algorithm had superior predicted performance for V-structure precision and recall and 2) no structural learning algorithm met our PCOR threshold or demonstrated acceptable V-structure recall performance. These insights were invaluable for decision making in the relevant context and showed us that we would need to further constrain our model or seek expert input before we could have confidence in our understanding of the effects of interventions on maternal health outcomes. The results also suggested that the FGES algorithm would provide the best model performance among those tested.  

\subsection{Causal Datasheet example: ALARM}

As a second proof of concept, we generated a Causal Datasheet for the ALARM Dataset(Appendix B). We approximated the characteristics of this dataset to mimic how a researcher might use the Causal Dataset Generation Tool. The ALARM dataset has 37 nodes with an average of 2.84 levels per node and a sample size of 20,000 \citep{beinlich1989alarm}. Using this information, we generated synthetic dataset experiments with similar characteristics (number of variables = 40, average number of levels = 3, and sample size = 20,000) and then generated a Causal Datasheet. The outputs from the CDG-T for the ALARM dataset gave us confidence that any of the tested algorithms could be used to achieve high skeleton precision and recall. We also learned that the FGES algorithm would likely have the highest performance related to V-structure recall and PCOR.  

\section{Discussion}
Having a causal datasheet that describes the expected causal BN performance in recovering ground truth structures for any given dataset tremendously valuable to both machine learning scientists and practitioners. It is clear from the generated datasheets that observed and unobserved dataset characteristics affect structural learning performance. These insights can help practitioners 1) during the experimental design phase, when they are interested in designing experiments with characteristics suitable for BN analysis, 2) during the analysis phase, when they are interested in choosing optimal structural learning algorithms and assigning confidence to DAGs, and 3) at the policy level, when they must justify their insights generated from BN analysis. We believe that this type of evaluation should be a vital component to a general causal inference work flow. 

While we believe the datasheet has utility in its current form, there are still a number of improvements to be made. We leave these as open questions to the community, and hope a collaborative effort can be made in order to further the usage of Bayesian networks in practice.
\setlist[enumerate]{
  labelsep=8pt,
  labelindent=0.5\parindent,
  itemindent=0pt,
  leftmargin=*,
  before=\setlength{\listparindent}{-\leftmargin},
}
\begin{enumerate}[label={\textbf{Question \arabic*}},nolistsep]
    \item How can we minimize the difference between the generated synthetic datasets and a real dataset to improve the accuracy of given estimates?
    \item How do we improve our coverage of the space of possible Bayesian networks, and are there other properties we should be capturing?
    \item How do we best characterize types of latent confounders, incorporate these into synthetic datasets, and measure their effects in order to estimate their effect in real datasets?
    \item Does the current cocktail of structure learning methods (MMHC, PC and, FGES) present a good coverage of the state of the art, and what hyper-parameter optimization methods should be used in order to ensure the best potential performance of each?
    \item What other metrics would be appropriate for recording the performance of structure learning, parameter estimation, and estimation of interventional odds ratios?
\end{enumerate}



 We were inspired by the problem of inferring causality from global development data sets and have estimated the range of data characteristics subjectively in that domain. Work is ongoing to incorporate the effects of latent confounders, continuous parameterization of structure types, missing data, noise, and added constraints. We would welcome further suggestions from theorists and practitioners alike.

\newpage
\bibliographystyle{ACM-Reference-Format}
\bibliography{NeurIPS_refs}
\pagebreak


\section*{Supplementary material}

# (Appendix A) Causal Datasheet for Maternal Health in Uttar Pradesh Dataset

> CDG-T provides overall recommendations for learning a BN with a given dataset. In this case, we learned that our dataset should *not* be used to learn a BN without further samples, constraints or inputs.

## Recommendations

The following are a list of recommendations for using Maternal Health in Uttar Pradesh Dataset to learn a Bayesian network:

- Given the **0.80 PCOR threshold** supplied by the user, the Maternal Health in Uttar Pradesh Dataset **should not** be used for learning a Bayesian network without further samples, constraints, or expert input.
- **All algorithms** can be used to obtain *perfect* (median 1) skeleton precision, and have high skeleton recall ($\geq 0.84$).
- **FGES** obtains perfect V-structure precision, but all algorithms have low V-structure recall ($\leq 0.54$).
- Given the low PCOR performance ($\leq 0.68$ PCOR) of all algorithms, the recommended strategy would be to use the **FGES** algorithm, using an expert to fill in missing edges while adding causal directions. An alternative to expert input would be to add constraints, or increase the number of samples.

> For a given set of dataset characteristics, CDG-T provides summaries of the performance of different structural learning algorithms for various aspects of BN model performance.

## Dataset Properties for Maternal Health in Uttar Pradesh Dataset

> Observable dataset characteristics are provided by the user into the CDG-T to look up algorithm performances and uncertainty estimations.

| Characteristic | Observable | Selected Values |
|---|---|---|
| Samples | ✓ | 6,000 |
| Variables | ✓ | 18 |
| Average Variable Levels | ✓ | 3 |
| Alpha (Imbalance) | - | 36 (Est.) |
| Complexity (Median Edges) | ✗ | Low, Med, High |
| Structure Type | ✗ | Forest Fire, Preferential Attachment, IC-DAG |

## Learning the Skeleton

The following are plots showing what skeleton structure learning performance to expect given Maternal Health in Uttar Pradesh Dataset's properties:

### Skeleton Precision

**Precision**, also known as the positive predictive value, is defined as the number of correct edge predictions made divided by all edge predictions made. The upper bound of precision is one, which corresponds to when all predicted true cases are correct. The lower bound is zero, which indicates that none of the predicted cases were correct.

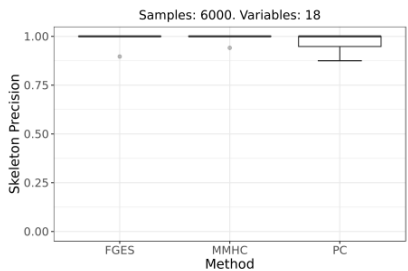

- Median precision using the **PC** algorithm = 1 (Q1= 1, Q3 = 0.95)
- Median precision using the **MMHC** algorithm = 1 (Q1= 1, Q3 = 1)
- Median precision using the **FGES** algorithm = 1 (Q1= 1, Q3 = 1)

### Skeleton Recall

**Recall** is the proportion of predicted positive cases over true positive cases. The upper bound of recall is one, corresponding to all true positive values being predicted. The lower bound is zero, indicating none of the true positive values have been captured.

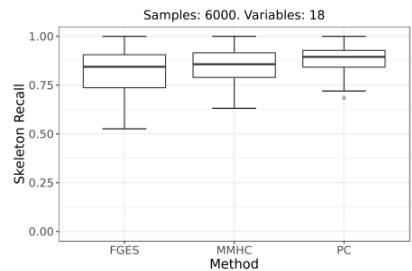

- Median recall using the **PC** algorithm = 0.9 (Q1= 0.93, Q3 = 0.84)
- Median recall using the **MMHC** algorithm = 0.86 (Q1= 0.92, Q3 = 0.79)
- Median recall using the **FGES** algorithm = 0.84 (Q1= 0.91, Q3 = 0.74)

## Learning the Direction

The following are plots showing what V-structure structure learning performance to expect given Maternal Health in Uttar Pradesh Dataset properties:

### V-Structure Precision

The precision with respect to V-Structures is the proportion of learnt V-Structures which are present in the true CPDAG.

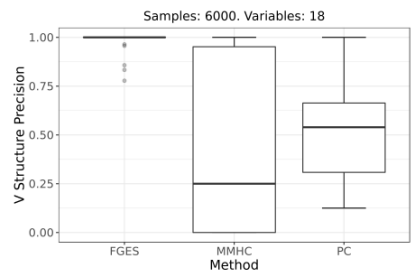

- Median V-Structure precision using the **PC** algorithm = 0.54 (Q1= 0.66, Q3 = 0.31)
- Median V-Structure precision using the **MMHC** algorithm = 0.25 (Q1= 0.95, Q3 = 0)
- Median V-Structure precision using the **FGES** algorithm = 1 (Q1= 1, Q3 = 1)

### V-Structure Recall

Recall of V-Structures is the proportion of the true V-Structures that are present in the predicted case.

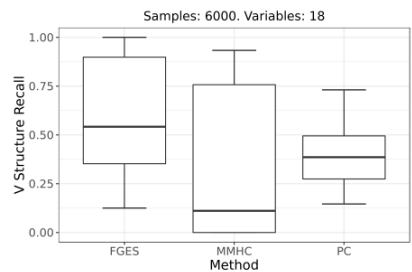

> CDG-T generates performance metrics and uncertainty estimated for different algorithms using a hypothesized range of potential ground truth properties. These metrics include skeleton precision and recall and V-structure precision and recall. A definition of each metric is provided to the user.

- Median V-Structure recall using the **PC** algorithm = 0.39 (Q1= 0.5, Q3 = .27)
- Median V-Structure recall using the **MMHC** algorithm = 0.11 (Q1= 0.76, Q3 = 0)
- Median V-Structure recall using the **FGES** algorithm = 0.54 (Q1= 0.9, Q3 = 0.35)

### Correctness of Causal Effects

The **Proportion of Correct Odds Ratios**(PCOR) measures the proportion of interventional odds ratios a learnt BN correctly estimates. This metric is calculated by first splitting odds ratios into three types of effects: protective (less than 1), detrimental (greater than 1), and neutral (uncertainty crosses 1). A matrix of these three effects is constructed for both the learned and true odds ratios, as show below.

| Learnt/True | Protective | Neutral | Detrimental |
|---|---|---|---|
| Protective | 1 | 0 | 0 |
| Neutral | 0 | 0 | 0 |
| Detrimental | 0 | 0 | 1 |

The above matrix is the one used in this evaluation. If the true effect is protective or detrimental, it is counted as *correct* when the learned odds ratio matches. Matching a neutral effect is not considered as a *correct* odds ratio, as we have found neutral effects are highly prevalent in balanced datasets. Neutral effects artificially inflate the results, and can create a false correlation.

> CDG-T provides the user with an estimate of PCOR. Depending on whether a practitioner is interested in outcomes versus interventions, they may choose to set their threshold for this value higher or lower. For example, if the usage case requires a relatively high confidence in the correct causal directions, the user may set a high PCOR threshold, e.g. PCOR > 0.80.

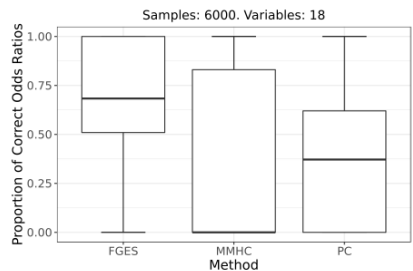

- Median PCOR using the **PC** algorithm = 0.37 (Q1= 0.62, Q3 = 0)
- Median PCOR using the **MMHC** algorithm = 0 (Q1= 0.83, Q3 = 0)
- Median PCOR using the **FGES** algorithm = 0.68 (Q1= 1, Q3 = 0.51)

### Susceptibility to Latent Confounders

The following plot shows performance degradation of the algorithms as latent confounders are introduced:

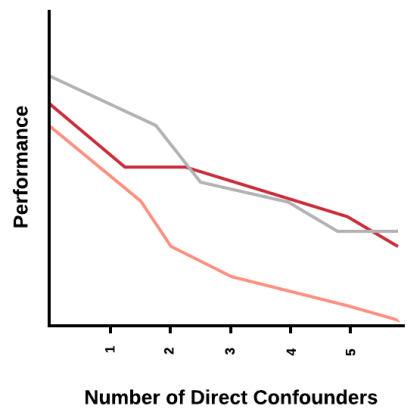

> Future iterations of CDG-T will provide users with insights into how potential missing latent confounders could affect model performance.

### Advantage of Constraints

Adding constraints, whether temporal or otherwise, can reduce errors when learning a Bayesian network. The following plot shows the advantage gained when grouping variables for the Maternal Health in Uttar Pradesh Dataset, with increasing granularity:

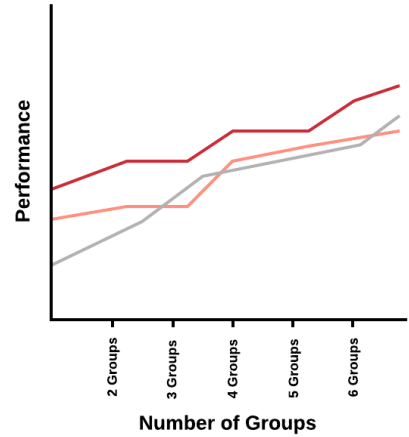

### Improving with More Samples

If the performance shown is this datasheet is not acceptable, improvements can generally be made by adding more samples. The following plot shows expected performance gain with further samples for Maternal Health in Uttar Pradesh Dataset:

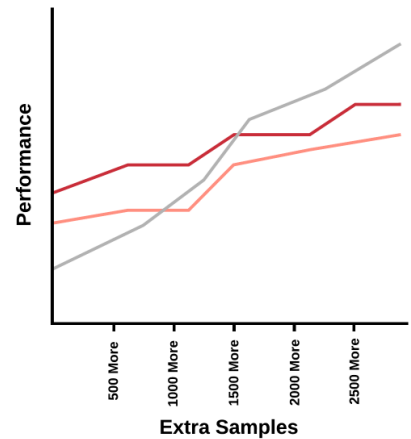

> Future iterations of CDG-T will help practitioners design experiments and analyses by providing information on how adding constraints or samples could improve model performance.

*Note: Susceptibility to Latent Confounder, Advantage of Constraints and Improving with Samples plots are current **placeholder** as this part of the CDG-T has not been completed.*

# (Appendix B) Causal Datasheet for ALARM Dataset

## Recommendations

The following are a list of recommendations for using ALARM Dataset to learn a Bayesian network:

- Given the **0.75 PCOR threshold** supplied by the user, the ALARM Dataset **can** be used for learning a Bayesian network.
- **All algorithms** can be used to obtain *high* ($\geq 0.98$) skeleton precision and have high skeleton recall ($\geq 0.91$)
- **FGES** obtains perfect V-structure precision, and good V-structure recall (median 0.85).
- The proportion of correct odds ratios is unacceptable for both **PC** and **MMHC** (median $\leq 0.5$); the PCOR for **FGES** exceeds the user threshold with a median of 0.85.

## Dataset Properties for ALARM Dataset

| Characteristic | Observable | Selected Values |
|---|---|---|
| Samples | ✓ | **20,000** |
| Variables | ✓ | **40** |
| Alpha (Imbalance) | ✗ | **6** (Est.) |
| Complexity (Median Edges) | ✗ | Low, Med, High |
| Structure Type | ✗ | Forest Fire, Preferential Attachment, IC-DAG |
| Average Variable Levels | ✓ | **3** |

## Learning the Skeleton

The following are plots showing what skeleton structure learning performance to expect given ALARM Dataset's properties:

### Skeleton Precision

**Precision**, also known as the positive predictive value, is defined as the number of correct edge predictions made divided by all edge predictions made. The upper bound of precision is one, which corresponds to when all predicted true cases are correct. The lower bound is zero, which indicates that none of the predicted cases were correct.

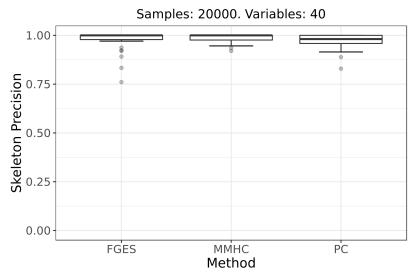

- Median precision using the **PC** algorithm = 0.98 (Q1= 1, Q3 = 0.96)
- Median precision using the **MMHC** algorithm = 1 (Q1= 1, Q3 = 0.98)
- Median precision using the **FGES** algorithm = 1 (Q1= 1, Q3 = 0.98)

### Skeleton Recall

**Recall** is the proportion of predicted positive cases over true positive cases. The upper bound of recall is one, corresponding to all true positive values being predicted. The lower bound is zero, indicating none of the true positive values have been captured.

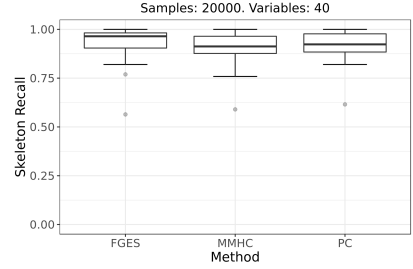

- Median recall using the **PC** algorithm = 0.92 (Q1= 0.98, Q3 = 0.88)
- Median recall using the **MMHC** algorithm = 0.91 (Q1= 0.97, Q3 = 0.88)
- Median recall using the **FGES** algorithm = 0.97 (Q1= 0.98, Q3 = 0.90)

## Learning the Direction

The following are plots showing what V-structure structure learning performance to expect given ALARM Dataset properties:

### V-Structure Precision

The precision with respect to V-Structures is the proportion of learnt V-Structures which are present in the true CPDAG.

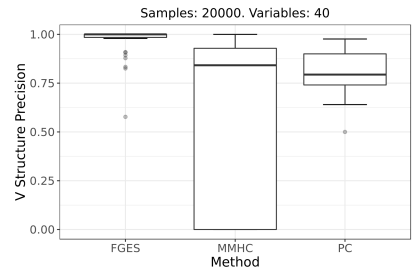

- Median V-Structure precision using the **PC** algorithm = 0.79 (Q1= 0.9, Q3 = 0.74)
- Median V-Structure precision using the **MMHC** algorithm = 0.84 (Q1= 0.93, Q3 = 0)
- Median V-Structure precision using the **FGES** algorithm = 1 (Q1= 1, Q3 = 0.98)

### V-Structure Recall

Recall of V-Structures is the proportion of the true V-Structures that are present in the predicted case.

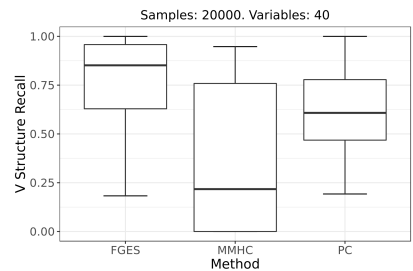

- Median V-Structure recall using the **PC** algorithm = 0.61 (Q1= 0.78, Q3 = .47)
- Median V-Structure recall using the **MMHC** algorithm = 0.22 (Q1= 0.76, Q3 = 0)
- Median V-Structure recall using the **FGES** algorithm = 0.85 (Q1= 0.96, Q3 = 0.63)

## Correctness of Causal Effects

The **Proportion of Correct Odds Ratios**(PCOR) measures the proportion of interventional odds ratios a learnt BN correctly estimates. This metric is calculated by first splitting odds ratios into three types of effects: protective (less than 1), detrimental (greater than 1), and neutral (uncertainty crosses 1). A matrix of these three effects is constructed for both the learned and true odds ratios, as show below.

| Learnt/True | Protective | Neutral | Detrimental |
|---|---|---|---|
| Protective | 1 | 0 | 0 |
| Neutral | 0 | 0 | 0 |
| Detrimental | 0 | 0 | 1 |

The above matrix is the one used in this evaluation. If the true effect is protective or detrimental, it is counted as *correct* when the learned odds ratio matches. Matching a neutral effect is not considered as a *correct* odds ratio, as we have found neutral effects are highly prevalent in balanced datasets. Neutral effects artificially inflate the results, and can create a false correlation.

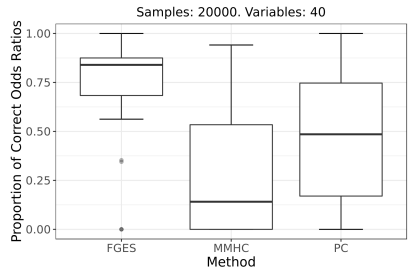

- Median PCOR using the **PC** algorithm = 0.46 (Q1= 0.75, Q3 = 0.17)
- Median PCOR using the **MMHC** algorithm = 0.14 (Q1= 0.53, Q3 = 0)
- Median PCOR using the **FGES** algorithm = 0.84 (Q1= 0.68, Q3 = 0.88)

## Susceptibility to Latent Confounders

The following plot shows performance degradation of the algorithms as latent confounders are introduced:

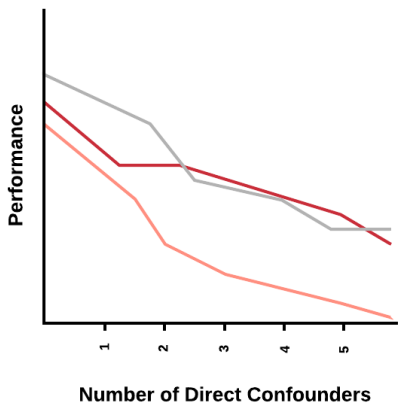

## Advantage of Constraints

Adding constraints, whether temporal or otherwise, can reduce errors when learning a Bayesian network. The following plot shows the advantage gained when grouping variables for the ALARM dataset, with increasing granularity:

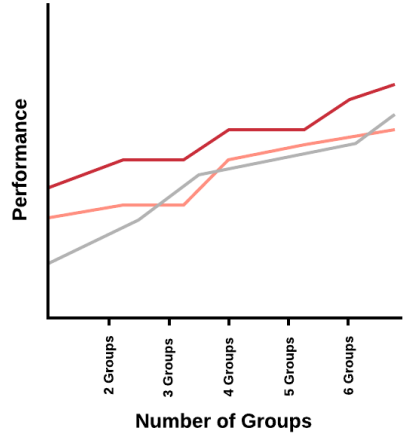

## Improving with More Samples

If the performance shown is this datasheet is not acceptable, improvements can generally be made by adding more samples. The following plot shows expected performance gain with further samples for ALARM Dataset:

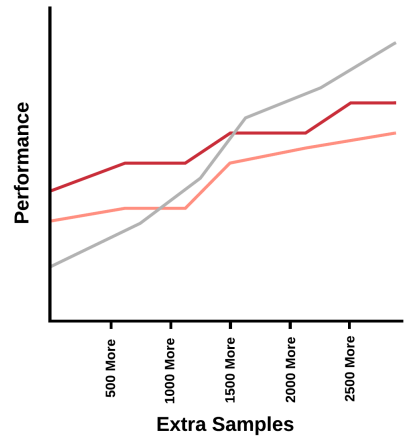

*Note: Susceptibility to Latent Confounder, Advantage of Constraints and Improving with Samples plots are current **placeholder** as this part of the CDG-T has not been completed.*



\end{document}